%% file: main.tex
\documentclass[preprint,10pt,a4paper]{elsarticle}

\usepackage{titlesec}
\usepackage[dvipsnames]{xcolor}

\usepackage[utf8]{inputenc}
\usepackage[english]{babel}
\usepackage[T1]{fontenc} 
\usepackage{verbatim}
\usepackage{dutchcal}
\usepackage{graphicx}
\usepackage{adjustbox}
\graphicspath{{Images/}} 
\usepackage{eso-pic} 
\usepackage{subfig} 
\usepackage{caption} 
\usepackage{transparent}

\usepackage{booktabs}

\usepackage{amsmath}
\usepackage{amsthm}
\usepackage{bm}
\usepackage[overload]{empheq}  

\usepackage{tabularx}
\usepackage{longtable} 
\usepackage{colortbl}

\usepackage{algorithm2e}

\usepackage[autostyle]{csquotes}
\MakeOuterQuote{"}

\usepackage{amssymb}
\usepackage{amsmath}

\usepackage[colorlinks=true,linkcolor=black,anchorcolor=black,citecolor=black,filecolor=black,menucolor=black,runcolor=black,urlcolor=black]{hyperref} 
\usepackage{cleveref}
\usepackage{tabularx}

\usepackage{appendix}

\usepackage{enumitem}

\usepackage{smartdiagram}

\usepackage{amsthm,thmtools,xcolor} 
\usepackage{comment} 
\usepackage{fancyhdr} 
\usepackage{lipsum} 
\usepackage{tcolorbox} 
\usepackage{xcolor}
\usepackage{stfloats} 
\usepackage{tikz}
\usepackage{pgfplots}

\usepgfplotslibrary{external}
\pgfplotsset{compat=1.14}

\newcommand{\approach}{ALIF\xspace}

\title{Active Learning-based Isolation Forest (\approach): Enhancing Anomaly Detection in Decision Support Systems}

\author[inst1]{Elisa Marcelli}       
\author[inst1]{Tommaso Barbariol}
\address[inst1]{Department of Information Engineering, University of Padova, Via Giovanni Gradenigo 6, Padova, 35131, PD, Italy}

\author[inst1,inst2]{Gian Antonio Susto}

 \address[inst2]{Human Inspired Technology Research Centre, University of Padova, Via Luigi Luzzatti, Padova, 35121, PD, Italy}        
\journal{Information Sciences}
\begin{document}

\begin{abstract} 
The detection of anomalous behaviours is an emerging need in many applications, particularly in contexts where security and reliability are critical aspects. While the definition of anomaly strictly depends on the domain framework, it is often impractical or too time consuming to obtain a fully labelled dataset. The use of unsupervised models to overcome the lack of labels often fails to catch domain specific anomalies as they rely on general definitions of outlier. This paper suggests a new active learning based approach, \approach, to solve this problem by reducing the number of required labels and tuning the detector towards the definition of anomaly provided by the user.
The proposed approach is particularly appealing in the presence of a Decision Support System (DSS), a case that is increasingly popular in real-world scenarios. While it is common that DSS embedded with anomaly detection capabilities rely on unsupervised models, they don't have a way to improve their performance: \approach is able to enhance the capabilities of DSS by exploiting the user feedback during common operations. 
\approach is a lightweight modification of the popular Isolation Forest that proved superior performances 
with respect to other state-of-art algorithms in a multitude of real anomaly detection datasets. 


\end{abstract}

\begin{keyword}
Active Learning \sep Anomaly Detection \sep Decision Support System \sep Industry 4.0 \sep Isolation Forest \sep Machine Learning \sep Weakly Supervised Learning
\end{keyword}
\maketitle




\input{introduction.tex}
\input{related_works.tex}
\input{model.tex}
\input{experiments.tex}
\input{conclusions.tex}

\section*{Acknowledgement}
This work has been supported by MIUR (Italian Minister for Education) under the initiative “Departments of Excellence” (Law 232/2016) and by "Black-box Anomaly Detection: Advanced Approaches and Applications - BADA$^3$" funded by the Department of Information Engineering of University of Padova.

\bibliography{bibliography.bib}

\end{document}

%% file: introduction.tex
\section{Introduction}
\label{intro}

Machine Learning (ML) gained a huge success in the last decades, becoming one of the most popular and studied branches of artificial intelligence \cite{jordan2015machine}. ML methods are widely used in many fields of research, with the aim of obtaining a general working learning rule from input data, namely a prediction function, to be used for future predictions of never-seen-before data.
Specifically, ML algorithms have been widely exploited in industrial processes, playing a relevant role in a wide range of applications: Industry 4.0 \cite{angelopoulos2019tacklin}, healthcare \cite{kourou2015machine}, transportation \cite{hamner2010predicting}, natural science \cite{yao2008quantitative}, social media \cite{balaji2021machine}, fraud detection \cite{awoyemi2017credit} and so on.

Anomaly detection represents an important and widely used ML task, broadly applied in various domains and applications where the issue of monitoring unexpected data behaviour is essential. This task defines the process of identifying anomalous data, i.e., data being characterised by a different behaviour with respect to other data distinguishing itself from the rest of the dataset. To date, there is no unanimously accepted definition but broadly speaking, an anomalous point, often named equivalently anomaly or outlier, is defined \textit{as an observation that deviates so much from other observations as to arouse suspicion that it was generated by a different mechanism} \cite{hawkins1980identification}. Anomaly detection is commonly tackled in many industrial scenarios, such as credit card fraud detection \cite{ghosh1994credit}, insurance fraud detection \cite{fawcett1999activity}, insider trading detection \cite{donoho2004early}, medical anomaly detection \cite{wong2003bayesian}. In these dynamic and often complex contexts, the problem of detecting anomalies is crucial in order to predict and avoid failures as well as to perform fault detection. In many industrial processes in fact, data-driven approaches for smart monitoring (for example predictive maintenance) have a key role, allowing to identify and isolate faults and to prevent future sudden failures. To solve this problem, anomaly detection represents an efficient solution.
Generally, in this scenario a great amount of collected data are available but, since labelling is an expensive and time consuming process, there is a lack of ground truth labels, undoubtedly stating whether or not a point is anomalous. The learning problem therefore is unsupervised and the algorithm can just blindly look at the structure of the dataset, without a clear definition of what is an anomaly from the user perspective. Therefore unsupervised algorithms can only detect samples that exhibit some general property different to the rest of the dataset, for example some approaches look for points far from the majority, or detect points living in low density areas.

Unfortunately, anomalies are strongly domain specific \cite{foorthuis2021nature}: as stated above, since no official definition is given, the concept of what an anomaly is entirely relies on the application in question. Specifically, it may happen that a set of data has different anomalies based on the given application domain and that the same data may be considered anomalous in one domain but normal in another \cite{sejr2021explainable}. For instance looking at data acquired by a measuring instrument, the manufacturer might define anomalies as events where the instrument has a faulty behaviour while the end-user might be more interested in events where the measured process behaves in a previously unseen way \cite{barbariol2020self}. As a direct consequence, training a domain specific anomaly detector would require a full set of labeled data to capture the user definition of anomaly. 

In real world applications, assigning labels to input data pose a considerable challenge to take into account \cite{zhu2009introduction}. In order to train reliable models, a large amount of labeled data is needed but, in practical scenarios, labeled examples are limited or often too expensive and time-consuming to collect, leading to a huge issue to face. Obtaining labels requires an often too expensive cost to take care of since the labeling procedure is usually carried on by a human domain expert who manually labels each point with a time-consuming and demanding routine. Moreover, by definition anomalous points are rare and difficult to spot, making the problem a difficult challenge to be solved. 

Due to the difficulty of finding labeled points, in practical contexts anomaly detection is often treated as an unsupervised learning task. For the classical unsupervised anomaly detection problem, the purpose is to detect outliers with no use of labeled data based on the fact that normal data greatly outnumbers anomalous data, and anomalies are very different with respect to inliers. Unsupervised anomaly detection models are not tuned for the precise domain of application but are generally based on identifying rules based on specific data characteristics, such as density based algorithms, distance based methods etc. \cite{hochenbaum2017automatic, hill2010anomaly, knorr1998algorithms, breunig2000lof}. 
However recent literature \cite{das2016incorporating,sejr2021explainable} distinguishes the outliers to the anomalies: the first are the points highlighted by an unsupervised model, while the second are the ones the user actually sees as anomalous. As the unsupervised model is not directly tuned to the detection of the anomalies, the outliers might weakly correlate with them. 

\begin{figure}
    \centering
    \includegraphics[scale=0.6]{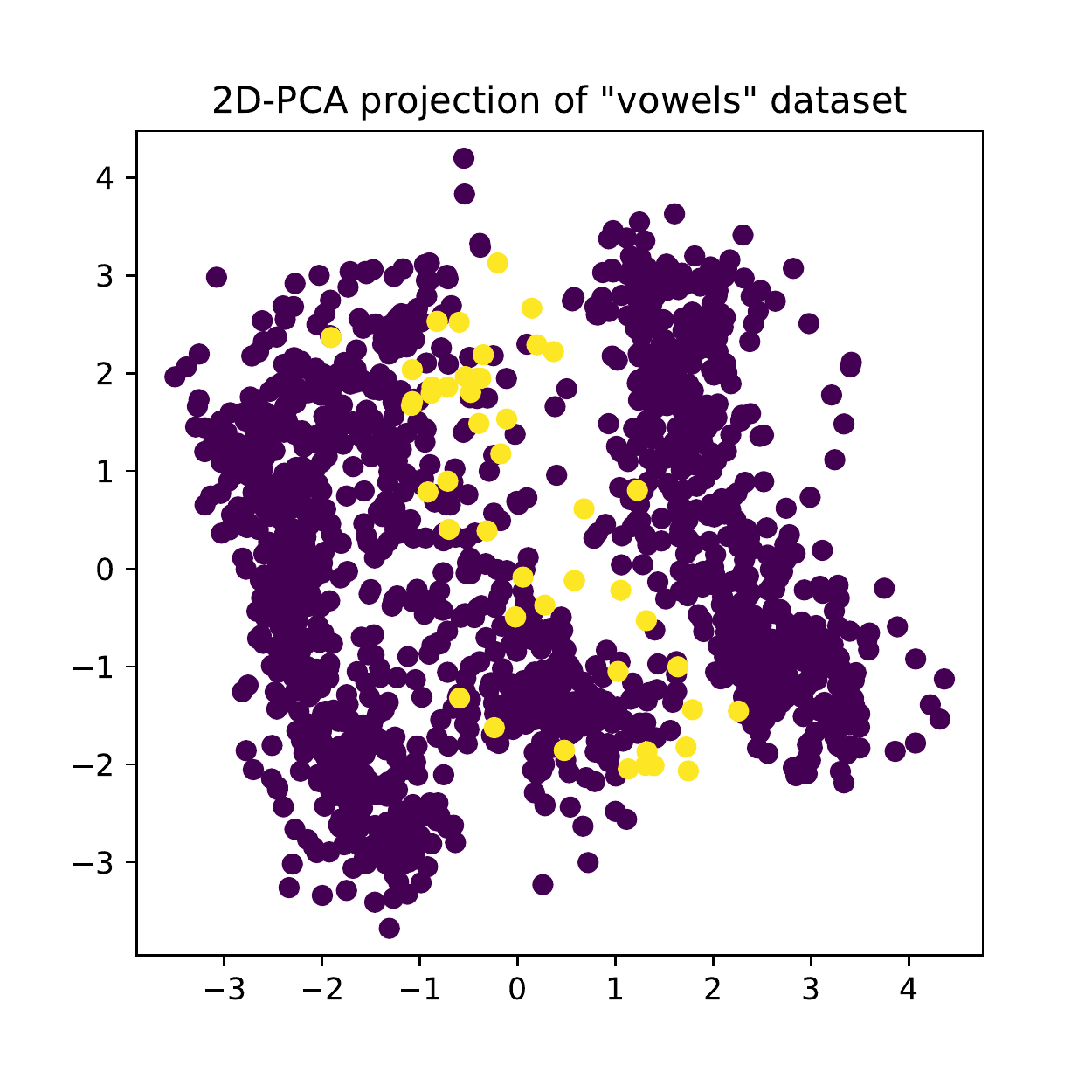}
    \caption{Projection of the \emph{vowels} dataset on 2 dimensions using Principal Component Analysis. Purple data points are normal data, yellow points are anomalous data. Two aspects are clearly visible: i) it is impossible to separate anomalies from normal data with only two features and ii) anomalies tend to form a different class and might be quite different from general outliers. Not all the data points lying in low density areas or far from the majority are defined as anomalies, but just the ones lying in a specific part of the space. 
   As a consequence, in the considered scenario, identifying anomalies might be a challenging task for an unsupervised detector.}
    \label{anomalyvsoutlier}
\end{figure}

Therefore, running unsupervised anomaly detection algorithms may be risky and often misleading: as stated above anomalies are strongly domain dependent and as a direct consequence, an unsupervised detector might not identify anomalous data which should be considered as such, as well as could wrongly detect as anomalies points which are normal based on the context taken under consideration \cite{das2016incorporating}. Figure \ref{anomalyvsoutlier} presents a visual representation of the strong connection between anomalous data points and context domain. Specifically, the plot shows the two-dimensional projection of the \textit{vowels} dataset \cite{Rayana}. As it can be seen, anomalies are not defined just as data points lying far or in low-density regions, but they form a class with a specific pattern defined by domain-experts, making complicated for the automatic detector to correctly identify them.

Among the unsupervised models, a very popular anomaly detection algorithm is the Isolation Forest (IF) \cite{liu2008isolation, liu2012isolation}, which presents a very different approach w.r.t. the majority of models: instead of creating a profile for normal data, it explicitly tries to isolate anomalies. To do it, IF relies on two assumptions: anomalies are fewer in number and they have very different attributes compared to normal data.


In Decision Support Systems (DSS) \cite{keen1980decision}, data streams are analysed in order to quickly extract strategic decisions on complex problems. Such process is monitored by users who frequently interact with the system and represent the actual decision maker of the whole process. In such framework, \approach represents an extremely appealing approach. Specifically, if a DSS is present, as a direct consequence, a user is already overseeing the process and inspecting data points: considering an unsupervised anomaly detection problem, inexpensive labels may be obtained in a fast way and using \approach the model may be inexpensively updated.

In this paper we describe a procedure able to tune the detector model on domain specific anomalies by interacting with a human expert. To perform the proposed tuning method, not every training data are presented and labeled but a subset is automatically selected so that the number of interactions between the system and the human is minimised. The core idea is to ask labels corresponding to the most significant points to reduce the labeling cost and at the same time to maximize the detection performance. As a direct consequence, the proposed procedure may be regarded as an Active Learning (AL) based model \cite{kumar2020active}. 

\input{tikz_AL}

Indeed AL represents a training approach particularly suitable when labeled samples are too expensive or difficult to obtain. Specifically, AL is a particular ML algorithm based on a key idea: despite the shortage of labeled data, high accuracy results may be obtained if the training algorithm is allowed to choose the points to be labeled and learn from them \cite{settles1995active}. An AL algorithm asks an oracle to label the data considered most informative with an iterative approach. Doing so, since the queried points are directly selected by the learning algorithm, the amount of necessary labeled data is much smaller than that required for classical supervised ML approach. Figure \ref{al} shows the core structure of any AL algorithm: at each iteration the model is updated using the labelled dataset, and is allowed to ask for a new label in the unlabelled dataset. This process repeats until the model reaches sufficient performances or when the number of iterations reaches the maximum budget.


This paper focuses on the Isolation Forest detector, and suggests a strategy to tune it towards the user definition of anomaly. In this work the authors compare two AL query policies to ask the user new labels, and other two policies to update the internal structure with minimal computational effort. The goal is to increase the performance of the detector as much as possible, keeping very low both the labelling effort and the updating procedure. Moreover this method has two key advantages over the supervised and computationally expensive models: as it relies on an initial unsupervised training, it can start to work when there are no labels, but more importantly it can work even if instances from only one class are labelled. This is particularly useful when obtaining labels from the anomalous class is very uncommon or expensive.

The rest of the paper is organized as follows. Initially, in Section \ref{rw} we outline the Isolation Forest in detail, and we indicate an existing active learning-based anomaly detection algorithm that will be used as a benchmark in thos work. Then, in Section \ref{pm} we illustrate the proposed model \approach: namely, we describe the strategies suggested to query the points as well as the approaches employed to update the model. In Section \ref{exp} we test \approach, comparing it with other models in relation to multiple real set of data. Finally, in Section \ref{conclusions} we draw conclusions for the present work.

\input{notation}

%% file: tikz_AL.tex
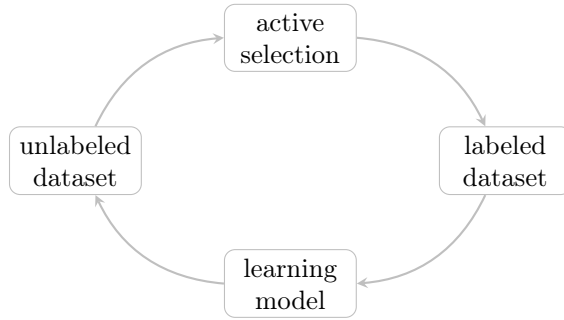
\begin{figure}
\centering
\usetikzlibrary{automata, arrows.meta, positioning}
 \begin{tikzpicture} [node distance = 4cm, on grid, auto]
 
\node (q0) [draw,lightgray, rounded corners, text width=1.5cm,yshift=-1.2cm, align =center] {\textcolor{black}{unlabeled \\dataset}};
\node (q1) [draw,lightgray, text width=1.5cm, rounded corners,above right = of q0,  yshift=-1.2cm,align =center] {\textcolor{black}{active \\selection}};
\node (q2) [draw,lightgray,rounded corners, text width=1.5cm, below right = of q1,yshift=1.2cm,align =center] {\textcolor{black}{labeled \\dataset}};
\node (q3) [draw,lightgray,rounded corners, text width=1.5cm, below left = of q2,yshift=1.2cm,align =center] {\textcolor{black}{learning \\model}};
 
\path [-stealth, thick]
    (q0) edge [lightgray, bend left]  (q1)
    (q1) edge [lightgray,bend left]  (q2)
    (q3) edge [lightgray,bend left]  (q0)
    (q2) edge [lightgray,bend left]  (q3);
\end{tikzpicture}
\caption{Active learning core structure. At each iteration a novel point is actively selected from the unlabeled set of data and the corresponding label is requested. Based on the received information, the model is modified.} \label{al}
\end{figure}

%% file: notation.tex
\begin{table*}[]
    \centering
    \begin{tabular}{ll}
    \hline
         \textbf{Symbol} & \textbf{Description} \\
         \hline
         $X$ & generic sample set \\
         $X'$ & generic sub-sample set ($|X'|= \psi$) \\
         $X^\mathcal{s}$ & set of labelled training points\\
         $X^\mathcal{u}$ & set of unlabelled training points \\
         $n_x$ & number of observations in $X$\\
         $n_\mathcal{s}$ & number of observations in $X^\mathcal{s}$\\
         $n_\mathcal{u}$ & number of observations in $X^\mathcal{u}$\\
         $x_j$ & $j-$th queried point\\
         $F$ & forest \\
         $T$ & tree \\
         $n_T$ & number of trees\\
         $L$ & leaf \\
         $n_L$ & number of leaves \\
         $L_\mathcal{p}$ & partition of $X$ made by $L$\\
         $L_\mathcal{h}$ & depth of $L$\\
         $L_\mathcal{i}$ &  number of normal points in $L$\\
         $L_\mathcal{o}$ & amount of anomalies contained in $L$ \\
         $a(x)$ & anomaly score of generic point $x$ \\
         $h^\mathcal{u}(x)$ & "unsupervised" path length of generic point $x$ \\ 
         $h^\mathcal{s}(x)$ & "supervised" path length of generic point $x$ \\ 
         $k(L)$ & color of leaf $L$\\
         $\lambda_T(x)$ & compute leaf containing $x$ with respect to tree $T$\\
         $H_{jt}$ & compute the path length of $x_j$ with respect to tree $T_t$\\
         $H \in \mathbb{R}^{n_T \times n_\mathcal{u}}$ & matrix of elements $H_{jt}$ \\
         \hline
    \end{tabular}
    \caption{List of symbols used.}
    \label{list_sym}
\end{table*}

%% file: related_works.tex
\section{Related works}
\label{rw}
Isolation Forest \cite{liu2008isolation, liu2012isolation} represents an innovative and original anomaly detection algorithm. The primary benefits of the algorithm, distinguishing itself with respect to other existing approaches, depend on its core structure. In the majority of AD models, normal points have a central role: data points that do not fit with the computed model, and which therefore break away from computed normal instance profile, are classified as anomalies. Instead of creating a profile for normal points, Isolation Forest focuses its structure on anomalous points, relying on the hypothesis that anomalies are fewer in number and different in attribute-values and therefore easier to be separated from the rest of the data  with the use of a small number of partitions. Isolation Forest is a non-parametric cost-effective ensemble method that revealed to be successful, often exceeding the results of more elaborate state of the art methods \cite{susto2017anomaly}. 

Similarly to Random Forest \cite{ho1995random}, Isolation Forest training phase constructs an ensemble of decision trees, also known as isolation trees (iTrees), relying on the fact that anomalies are easier to be isolated, i.e., partitioned from the rest of the data, due to their distinctive features. First, Isolation Forest randomly sub-samples the dataset so that each iTree is obtained with a different set of data. Then, the structure of an iTree is generated in a completely random way: each partition is produced with a random selection of an attribute value from the subset at disposal and by the choice of the split value, selected randomly in the range of the picked attribute. This recursive procedure is repeated until all points are isolated or a predefined limit height is reached. After that, a novel random sub-sample is selected and the isolation procedure is repeated, in order to create a new random iTree. Once the training phase is completed and every iTree is fully grown, data traverse the different iTrees and its depths, i.e., the number of edges traversed from the root node to the external node containing it, are collected. Based on these depths $h^u$, also known as path lengths, the anomaly score is computed, i.e., an indicator of the likelihood that a point is an anomaly. 

Let $X$ be a generic set of data. Every iTree is generated from a sub-sample $X' \subset X$, $|X'|= \psi$: the recursive branching process starts by selecting a random feature and a random value between the minimum and the maximum value in that feature. At each branch, points are split into two daughter nodes based on the corresponding feature attribute. 
Let $x \in X$, then the corresponding anomaly score is defined as
\begin{equation}
    a(x) = 2^{-\frac{E(h^\mathcal{u}(x))}{c(\psi)}} \label{as}
\end{equation}
where $E(h^\mathcal{u}(x))$ is the average path length of $x$ with respect to all trees and $c(\psi)$ is a normalization factor with the sub-sample set size as input. Specifically, \[c(\psi)=2H(\psi - 1)-2(\psi-1)/\psi\] and defines the average path length of an unsuccessful search in a binary search tree computed with a set of cardinality $\psi$. Based on Eq. (\ref{as}): when $E(h^\mathcal{u}) \rightarrow \psi -1$, $a(x)\rightarrow 0$ and it is quite safe to consider $x$ a normal point; on the contrary when $E(h^\mathcal{u}) \rightarrow 0$, $a(x)\rightarrow 1$ and $x$ is most likely an anomaly; when $E(h^\mathcal{u}) \rightarrow c(\psi)$, namely when the average path length of $x$ is close to the normalization factor, then $a(x) \approx 0.5$ and the sample has no recognizable anomalous factor. Note that, in order to classify whether or not a point is anomalous, it is necessary to define a score based border value, establishing a division between anomalous points and normal ones. However, the choice of such border value is strongly data dependent, relying upon its target subject, making it not a trivial task to be managed \cite{hofmockel2018isolation}.

The proposed algorithm starts by growing a standard Isolation Forest.
After that, an active learning based approach is carried on: in an iterative way, the model is allowed to ask for a point to be labeled, obtaining the true information about its nature as a direct consequence. Based on it, the inner structure of each tree is modified to exploit the incoming information and to improve/calibrate the model if necessary.

A similar active learning anomaly detection algorithm using an optimization based approach is presented in \cite{das2016incorporating}. The paper describes an Active Anomaly Discovery (AAD) method where the points having the highest anomaly score are presented to an expert analyst in an iterative way, requesting the corresponding true label. Based on the information received, the algorithm tries to place the points labeled as anomalies as close as possible to the higher part of the model. Specifically, the model considers the structure defined by the Isolation Forest and associates to each leaf a weight value, starting from a uniform fixed value and, based on the feedback received, such weights are iteratively updated. Such updating approach is performed with the use of an iterative optimization problem, where the optimal weights are computed so that every labeled anomaly has a score which is higher with respect to the labeled normal points ones. Note that, in this approach the partitions computed by the Isolation Forest are not modified, only the corresponding weights are. The proposed method is applied into a tree-based detector \cite{das2017incorporating}, where the described updating approach is used into the Isolation Forest algorithm. Such algorithm is called IF-AAD. 

Here we present a novel approach, called Active Learning-based Isolation Forest (\approach), which differs from the aforementioned works because of its easy yet efficient formulation: independently from the size of the input set, the algorithm will execute in constant time, modifying the average path length of the input points but leaving unchanged the Isolation Forest partitions, with the perk of using both current and past information. Differently from IF-AAD and Random Forest, the proposed approach does not need to pass through an optimization procedure, leading to a much lighter update.  In Section \ref{exp} we report the performance benchmarking of our approach, IF-AAD and Random Forests.

%% file: model.tex
\section{Proposed model: \approach}
\label{pm}
In this paper we present an adaptive anomaly detection model for fixing leaf depths according to labels received from domain experts.
The core idea of the proposed approach is presented in Algorithm \ref{alg_albif} and relies on developing an Isolation Forest based model in which the detector has the possibility to query domain experts for labels. In this iterative environment, once the Isolation Forest is fully grown, the algorithm can choose the points to be labeled and, based on the novel information achieved, its core structure is modified, in order to obtain a strong increase in the performance, with the use of only a limited number of labeled points. For every iteration, such modification only takes place in the external node containing the queried point, maintaining the main structure of the Isolation Forest untouched. In this way, the main goal is, given the novel information, to update the Isolation Forest structure, readjusting it based on the labeled points.

\input{algorithm2}

Specifically, the proposed approach may be outlined as follows. Let $X=\{x^1,\dots, x^n\}$ be a generic training dataset, where $x^i \in \mathbb{R}^m$, $i=1,\dots,n$. First, the Isolation Forest algorithm is trained, leading to a forest $F=\{T_t\}_{t=1}^{n_T}$ of fixed number $n_T$ of fully grown iTrees. By construction, $T=\{L_l\}_{l=1}^{n_L}$ namely each tree $T$ is characterized by a variable number of leaves $L$. We use the following form to describe each leaf $L$:
$L=(L_\mathcal{p}, L_H, L_\mathcal{i}, L_\mathcal{o})$
where 
\begin{itemize}
    \item $L_\mathcal{p}$ defines the partition of $X$ made by $L$, describing where a leaf is located with respect to the input space;
    \item $L_\mathcal{h}$ is the depth of $L$;
    \item $L_\mathcal{i}$ refers to the number of normal points in $L$;
    \item $L_\mathcal{o}$ specifies the amount of anomalies contained in $L$.
\end{itemize}
As a direct consequence, initially each data $x \in X$ is assigned with the corresponding average path length $E(h^\mathcal{u}(x))$ and anomaly score $a_\psi(x)$ as computed by the Isolation Forest. 

From this step, the proposed iterative active learning approach can start. At each iteration, a point $x^{\mathcal{s}}$ is selected and the corresponding true label $y_{\mathcal{s}}$ is requested. Accordingly, each iTree $T$ is investigated, determining the leaf where the queried point lies, and modified as a result. We define $X^\mathcal{u}=\{x^{\mathcal{u}}\}$, $X^\mathcal{s}=\{x^{\mathcal{s}}\}$ and $Y^\mathcal{s}=\{y^{\mathcal{s}}\}$ respectively the set of unlabeled training points, the set of labeled inputs and the set of corresponding labels. Note that, by definition we have that $X= X^\mathcal{u} \cup X^\mathcal{s}$. 

The key intuition behind the modification of the structure of each iTree is very simple. First, a queried point is selected and the corresponding trusted label is obtained. Secondly, for each $T$ of the model, the external node containing the point is analysed: its depth is updated based on the achieved information so that true anomalies are located closer to the root node while true normal points are far away from it. 

\input{tikz_tree}

Based on this scenario, it is important to define two essential yet independent tasks on which the entire algorithm relies on:
\begin{enumerate}
    \item[i.] \textit{Update strategy:} The actual approach employed to modify the classical Isolation Forest model;
    \item[ii.] \textit{Query strategy:} The plan of action to choose the optimal method to select the queried points. 
\end{enumerate}
Both tasks are detailed in the following.

\subsection{Update strategy}
Let $L$ be the leaf under investigation. Then, based on the labeled points contained in it, i.e., based on the proportion of anomalies and normal points contained in $L$, we define the \textit{color} of $L$ 
\begin{equation}
    k(L) = \frac{L_\mathcal{o}-L_\mathcal{i}}{L_\mathcal{o}+L_\mathcal{i}},
    \label{color}
\end{equation}
where $L_\mathcal{i}$ and $L_\mathcal{o}$ are respectively the number of labeled normal points and the number of labeled anomalies sampled in the leaf. Therefore, the color of a leaf defines its inner structure, describing how the total amount of labeled data contained in it is distributed. Specifically, it outlines the probability of a leaf to be anomalous with respect to the labeled data in it. Based on the color of a leaf, the corresponding fixing procedure is performed. Figure \ref{strat_fig} shows how the \textit{color} of a leaf is computed in a visual way. 

Every time a novel point is queried, the model is updated based on the received information. In relation to each iTree, the update exclusively takes place with respect to the leaf containing the queried point. Specifically, let us consider a generic iTree $T$: once the queried point $x^{\mathcal{s}}$ is selected and the true label is obtained, $T$ is investigated and the leaf containing $x^{\mathcal{s}}$ is considered. Its depth $h^\mathcal{u}(x)$ (computed by the Isolation Forest) is forgotten and substituted with a supervised value $h^\mathcal{s}(x)$, entirely depending on the supervised data contained in the leaf, i.e., on $L_\mathcal{i}$ and $L_\mathcal{o}$. Algorithm \ref{alg_leafDepth} summarizes the above described procedure. 


Equations (\ref{color}) is used to obtain a leaf coefficient value $\bar{k}(L)$, describing the structure of the leaf with respect to the current labeled point as well as taking into account the past information received.
Specifically, based on the information in use, the corresponding leaf value becomes
\begin{equation}
    \bar{k}(L)= \frac{1}{2} \big(k+1 \big). \label{scol}
\end{equation}
For consistency with the rest of the manuscript we renamed the function in (\ref{scol}) as $k$. 
Note that, using Equation (\ref{scol}), the codomain of function $k(L)$ is given by the closed interval $[0, \, 1]$. Specifically, the following evaluations are made:
\begin{enumerate}
    \item[$\cdot$] If $L$ contains only normal points and it is fully labeled, then $k(L) \rightarrow 0$;
    \item[$\cdot$] If $L$ contains only anomalies and it is fully labeled, then $k(L) \rightarrow 1$; 
    \item[$\cdot$] If $L$ has a balanced number of both anomalies and normal points then $k(L) = \frac{1}{2}$.
\end{enumerate}
Obviously $k(L)$ is computed only for leaves containing some labelled data while the algorithm keeps untouched the leaves that are not actively sampled.

The novel supervised depth $h^\mathcal{s}(k)$ takes Equation (\ref{scol}) as argument. Specifically, we define two different approaches to compute $h^\mathcal{s}(k)$ defined as:
\begin{enumerate}
    \item \textit{Piece-wise Linear} supervised depth
    \begin{equation}
    \begin{split}
    h^\mathcal{s}(k)=\begin{cases}
     2k\big[c(\psi)-h_{max}\big]+h_{max},  \hspace{0.4cm} & \text{if} \;\; 0 \leq k < \frac{1}{2} \\
     & \\
     2k\big[h_{min}-c(\psi)\big]+2c(\psi)-h_{min}, \hspace{0.4cm} &\text{if} \;\; \frac{1}{2} \leq k \leq 1\\
    \end{cases} \label{syn}
    \end{split}
    \end{equation}

where $h_{max}$ and $h_{min}$ are respectively the minimum depth and the maximum depth of the Isolation Forest computed during the unsupervised training. 
    \item \textit{Logarithmic} supervised depth
    \begin{equation}
        h^\mathcal{s}(k)= - c(\psi) \log_2(k).
        \label{synlog}
    \end{equation}
\end{enumerate}
Note that, in Equations (\ref{syn})
\begin{enumerate}
    \item[$\cdot$] when $k(L) \rightarrow 0$, then $h^\mathcal{s}(k) \rightarrow h_{max}$;
    \item[$\cdot$] when $k(L) \rightarrow 1$, then $h^\mathcal{s}(k) \rightarrow h_{min}$;
     \item[$\cdot$] when $k(L) = \frac{1}{2}$, then $h^\mathcal{s}(k) = c(\psi)$.
\end{enumerate}
Regarding the logarithmic depth described in Equation (\ref{synlog}), when $k(L) = \frac{1}{2}$, then $h^\mathcal{s}(k) = c(\psi)$. Nevertheless, when $k(L) \rightarrow 1$, then $h^\mathcal{s}(k)$ converges to $0$ and, analogously, $h^\mathcal{s}(k) \rightarrow +\infty$ when $k(L) \rightarrow 0$. Unfortunately, this makes the logarithmic choice quite unstable when normal points are labelled. However, it is possible to apply a threshold on the logarithm to saturate it, leading to a behaviour very similar to the piece-wise linear function.
Figure \ref{syngraph} plots the relation connecting the leaf value $k(L)$ with the supervised depth $h^\mathcal{s}(k)$ for both the piece-wise linear depth and the logarithmic depth. 

It is important to note that, the proposed update strategy does not require to fully retrain the forest but, on the contrary, when novel information is acquired, \approach only modifies the actively sampled leaves, leaving the rest  unchanged. Therefore, the time complexity of the update strategy is linear with respect to the number $n_T$ of trees in the Forest.


\begin{figure}
\centering
\input{tikz_graph}
\caption{Visual representation of the two possible definitions of supervised depth. The blue function represents the piece-wise linear depth: when the leaf value $k(L)$ is close to 0, i.e., the leaf is fully labeled with normal points, the novel supervised depth of $L$ is tends to the maximum depth; if $k(L)$ takes a value close to 1, the leaf will be push to the minimum depth, as this situation corresponds to $L$ fully labeled of anomalies. The red function shows the logarithmic depth: in the range extremes $0$ and $1$, a threshold value is applied and the depth values are forcibly set to respectively $h_{max}$ and $h_{min}$.} \label{syngraph}
\end{figure}

\subsection{Query strategy}
The idea of incorporating expert feedback in unsupervised anomaly detection algorithms aims at improving the achieved performance adding a relatively small computational and labelling cost. 
To improve the model in an optimal manner, the choice of the proper strategy to select the points to be queried must be established. 
Beyond the employed strategy to modify the structure of the Isolation Forest based on the novel information in fact, the proposed model is significantly based on the choice of the queried points. Specifically, for the successful outcome of the method, choosing for the most appropriate point to be labeled is a key factor which could compromise its outcome. 
In classical active learning scenarios, several possible query strategies are presented \cite{settles1995active}. 

For any iTree $T_t \in F$ and input point $x_j \in X^\mathcal{u}$, let $\lambda$ be defined as
\begin{equation}
    \lambda_{T_t}(x) = L, \label{lambda}
\end{equation}
namely, function $\lambda$ assigns at each input point $x_j$ the leaf containing it with respect to tree $T_t$. Now, using Equation (\ref{lambda}), we define
\[H_{jt}\coloneqq h^\mathcal{s}(k(\lambda_{T_t}(x_j))) = h^\mathcal{s}(k(L)).\] 
Then, let $H \in \mathbb{R}^{n_T \times n_\mathcal{u}}$ be the following matrix 
\begin{equation*}
H = 
\begin{bmatrix}
H_{11} & \cdots & H_{1n_\mathcal{u}}\\
\vdots  &  \ddots & \vdots  \\
H_{n_T 1} & \cdots & H_{n_T n_\mathcal{u}}
\end{bmatrix}.
\end{equation*}
The choice of the point to request fully relies on matrix $H$.  
Details of the design of matrix $H$ can be found in Algorithm \ref{alg_DepthMatrix}.

Based on $H$, we define the two following query strategies:
\begin{enumerate}
    \item \emph{Maximum uncertainty}: at each iteration the point selected to be labeled $x^{\mathcal{s}}$ is the one where the iTrees disagree the most, namely
    \begin{equation}
        x^{\mathcal{s}}=\underset{j =1, \dots,  n_\mathcal{u}}{\operatorname{argmax}} \; \underset{t=1, \dots, n_T}{\text{std}} \; H \label{qs1}
    \end{equation}
    By doing so, the intended purpose would be to give the greater assistance to the model. 
   Specifically, at each iteration we ask for the label of the data point where the model is more unsure, adding information with respect to the most uncertain data.
    \item \emph{Most anomalous}: at each iteration the point selected to be labeled $x^{\mathcal{s}}$ is the point having the highest anomaly score value in the current iteration, namely
    \begin{equation}
        x^{\mathcal{s}}=\underset{j =1, \dots,  n_\mathcal{u}}{\operatorname{argmin}} \; \underset{t=1, \dots, n_T}{\text{mean}} \; H \label{qs2}
    \end{equation}
    This query strategy is the less expensive and more straightforward one and involves asking for the label of the point regarded as the most anomalous.
\end{enumerate}
Note that, in order to make the proposed approach feasible, for each of the listed strategies, each point may only be queried once.  When a point is queried and the corresponding label is obtained in fact, there is no need to ask for its label again since we are considering the information received undoubtedly true.

From a business process point of view, querying the most anomalous point may represent a sort of DSS, providing assistance to the decision making process and at the same time extracting information from a significant amount of data.  As stated above, a DSS software gathers data considered the most informative and, based on the information achieved, it generates analysis tools useful for the decision making process. 
\\In this scenario when a point is strongly considered an anomaly from the system, the domain expert is trigged for a checking. In this situation, the expert attention is drawn: the point has to be analysed in order to provide the actual corresponding label. Doing so, novel data information is obtained, combining the expert's knowledge together with the computerized system. In this way, the expert may validate the decision-making process and at the same time quickly extract useful information for the decision-making process. 

From a model based viewpoint, asking for the point where the iTrees are more uncertain may be the most relevant strategy. With this approach in fact, the assistance provided by the expert aims at addressing the decisions where the model struggles the most and, doing so, at updating the most critical situations. Of course, given the unbalanced number of anomalies, it is highly likely that, using the maximum uncertainty strategy, the vast majority of the labeled points would be normal data, making it possibly less suitable in a more practical prospective. 

As specified by Equations (\ref{qs1}) and (\ref{qs2}), \approach query strategy corresponds to searching along the number of rows and columns of matrix $H$ and has time complexity $O(n_X n_T)$.

%% file: algorithm2.tex
\begin{algorithm}
\caption{\textit{\approach$((x^{\mathcal{s}},y^{\mathcal{s}}), X^{\mathcal{u}}, F)$}}\label{alg_albif}
\KwData{labeled point $(x^{\mathcal{s}},y^{\mathcal{s}})$, unlabeled dataset $X^{\mathcal{u}}$, forest $F$}
\KwResult{updated forest $F$, query point $x^{\mathcal{s}}$} 
 $F \gets$ \textit{LeafDepthUpdate}$((x^{\mathcal{s}},y^{\mathcal{s}})$, $F)$\;
     $H \gets$ \textit{GetDepthMatrix}$(X^{\mathcal{u}}$, $F)$\;
    $x^{\mathcal{s}} \gets$ \textit{GetQuery}$(H)$

\end{algorithm}

\begin{algorithm}
\caption{\textit{LeafDepthUpdate$((x^{\mathcal{s}},y^{\mathcal{s}})$, $F)$}}\label{alg_leafDepth}
\KwData{labeled point $(x^{\mathcal{s}},y^{\mathcal{s}})$, unlabeled dataset $X^{\mathcal{u}}$, forest $F$}
\KwResult{updated leaf $L$} 
\For{$T_t$ \text{in} $F$}{
        $L \gets \lambda_{T_t}(x^{\mathcal{s}})$ \;
        \If{$y_{\mathcal{s}}==$ \text{anomaly}}{$L_\mathcal{o} \gets L_\mathcal{o} + 1$ \;}
        \Else{$L_\mathcal{i} \gets L_\mathcal{i} + 1$ \;}
        $L_\mathcal{h} \gets  h^\mathcal{s}(k(L))$
}
\end{algorithm}

\begin{algorithm}
\caption{\textit{GetDepthMatrix$(X^{\mathcal{u}}, F)$}}\label{alg_DepthMatrix}
\KwData{unlabeled dataset $X^{\mathcal{u}}$, forest $F$}
\KwResult{updated leaf $L$} 
\For{$T_t$ \text{in} $F$}{
    \For{$x^{\mathcal{u}}_j$ \text{in} $X^{\mathcal{u}}$}{
            $L \gets \lambda_{T_t}(x^{\mathcal{u}}_j)$ \;
            $H_{jt} \gets  h^\mathcal{s}(k(L))$
    }
}
\end{algorithm}

%% file: tikz_tree.tex
\begin{figure}
\usetikzlibrary{matrix, shapes.geometric}
\tikzstyle{dotted} = [circle, minimum width=3pt,fill, inner             sep=0pt]
    \centering
      \begin{tikzpicture} [scale=0.8]
        [square/.style={regular polygon, regular polygon sides=4}]
            \node[dotted] at (0,0)  (a) {};
            \node[dotted] at (3,-1) [circle,draw] (c) {};
             \node[dotted] at (4,-2) [circle,draw] (c1) {}; 
             \node[dotted,label={[yshift=0.1cm]:$L$}] at (2,-2) [circle,draw] (c2) {};
            \node[dotted] at (-3,-1) [circle,draw] (b) {};
            \node[dotted] at (-4,-2) [circle,draw] (b1) {};
            \node[dotted] at (-5,-3) [circle,draw] (b2) {};
            \node[dotted] at (-2,-2) [circle,draw] (b3) {};
            \node[dotted] at (-3,-3) [circle,draw] (b4) {};
            \draw (a) -- (c);
            \draw (a) -- (b);
            \draw (b) -- (b1);
            \draw (b1) -- (b2);
            \draw (b) -- (b3);
            \draw (b1) -- (b4);
            \draw (c) -- (c1);
            \draw (c) -- (c2);
            \node[circle,draw,NavyBlue, thick,dashed, minimum size=.3cm] (cir1) at  (2,-2) {};
            \node[circle,draw,NavyBlue, thick, dashed, minimum size=1.7cm] (cir2) at  (1,-3) {};
               
            \draw[NavyBlue, thick] (cir1) -- (cir2);
                
            \matrix(A)at (1,-3)[matrix of nodes,  rounded corners,
                nodes in empty cells, nodes={circle,scale=0.3, fill=gray, minimum size=5mm},
                column sep=1.5mm, row sep=1.5mm]
                { &|[fill=Green]|&\\
                  |[fill=RedOrange]|&&|[RedOrange]|\\
                 &|[fill=Green]|&|[Green]|\\
                  };

    \node[text width=6cm, anchor=west, right] at (3.4,-2){
    \begin{equation*}
k(L) = \frac{1}{2} \bigg(\frac{\color{RedOrange}{2} \  \color{Black}{-}  \ \color{Green}{3}}{\color{RedOrange}{2} \  \color{Black}{+}  \ \color{Green}{3}} + 1\bigg)
\end{equation*}};

\end{tikzpicture}

\caption{Every time a novel point is queried and the corresponding label is obtained, every iTree is investigated and the leaves containing the queried point are considered. A visual representation of the color of a generic leaf $L$ is displayed: green dots represents normal labeled instances; labeled anomalies are depicted by red dots. The corresponding color $k(L)$ is computed using Equation (\ref{scol}).} \label{strat_fig}
\end{figure}

%% file: tikz_graph.tex
\begin{tikzpicture} 
\begin{axis}[axis line style={->},xlabel=$k(L)$,axis y line=middle, x label style={at={(1.,0)},anchor=north},ylabel=$h^\mathcal{s}(k)$,y label style={at={(-10,36)},anchor=left}, compat=newest,
xmin=0,xmax=1.15,ymin=0,ymax=38, axis lines=center, ytick={1,7.5,10}, yticklabels={$h_{min}$,$c(\psi)$,$h_{max}$}, xtick={0.5,1}, xticklabels={$\frac{1}{2}$,1}]

\addplot[domain=0:1, color=red, thick]{-7. 5*log2(x)};
\addplot[domain=0:0.5, color=blue, thick]{-5*x+10} ;
\addplot[domain=0.5:1, color=blue, thick]{-13*x+14} ;
\draw [dashed,help lines] (0,1) -| (1,0);
\draw [dashed,help lines] (0,7.5) -| (0.5,0);
\end{axis}

\node[] at (0,-0.275){$0$};
\end{tikzpicture}

%% file: experiments.tex
\section{Experimental Results}
\label{exp}

\begin{figure}
    \centering
    \includegraphics[width=\textwidth]{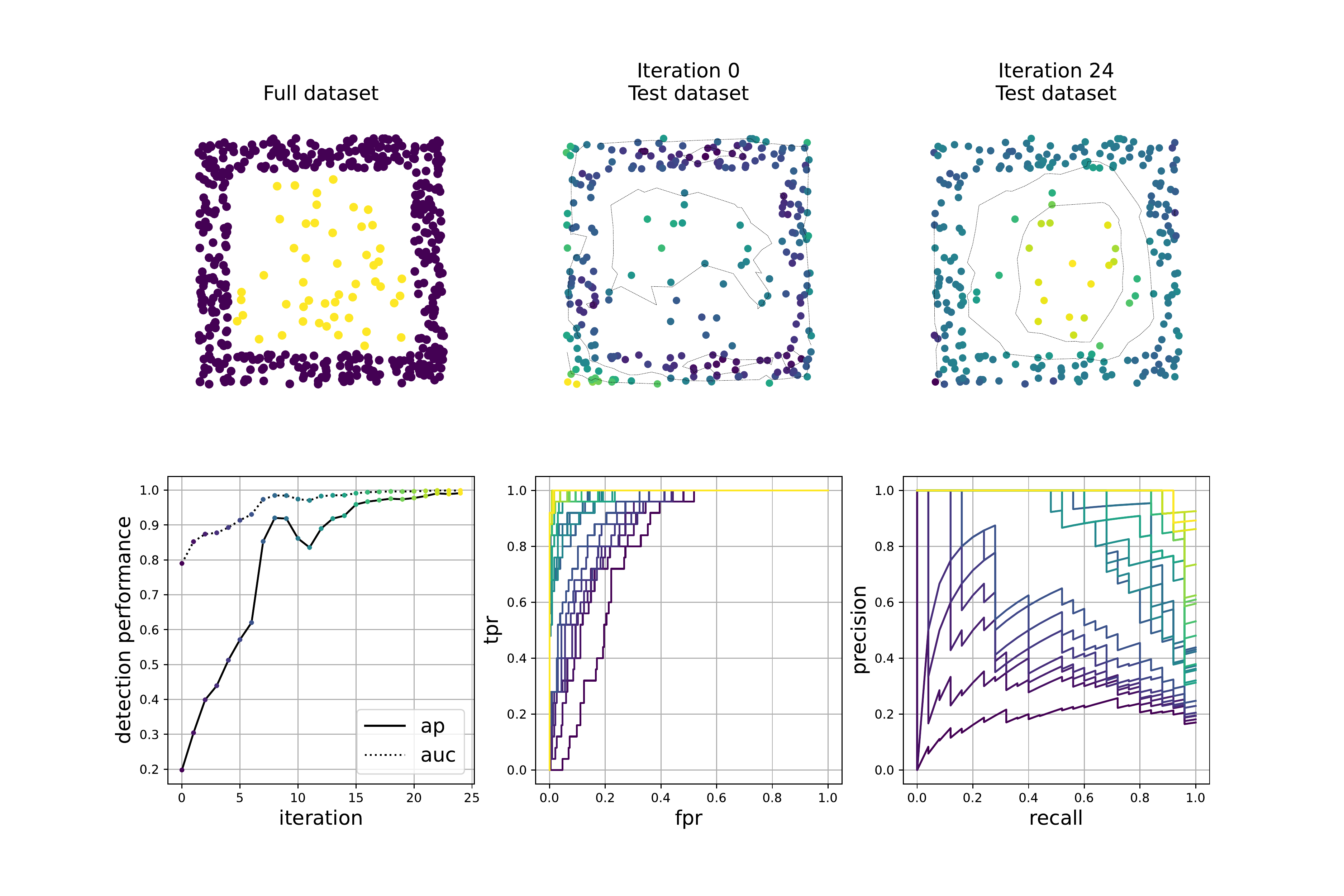}
    \caption{Test of the algorithm on a square toroidal dataset: anomalies lie inside of a box made up of normal instances. This setting is particularly challenging for the Isolation Forest algorithm as it is much easier to quickly separate normal points (depicted in purple) w.r.t anomalies (yellow). The anomaly score on test samples is shown on the second and third panel: the yellow is assigned to the points having the highest anomaly score, the purple viceversa. In the second row of panels the performances of the detector at each iteration is depicted: the \emph{area under the ROC curve} (auc) and the \emph{average precision} (ap) measured on the test set quickly improve. }
    \label{fig:toroid}
\end{figure}


In this section,  we analyse the performance of \approach, matching both the proposed update strategies with the two described query approaches. Doing so, we hope to obtain a full and detailed evaluation of the presented model, with the purpose of analysing the efficiency of each combination as well as giving meaningful guidance on the proper use of the proposed strategies.


Firstly, we tested our approach on synthetic datasets like the challenging shape depicted in Figure \ref{fig:toroid}, where the normal data make up a square toroid and the anomalous data lie inside it. In this kind of datasets, the Isolation Forest perform quite badly and there is a lot of room for improvement as it struggles to  separate in few steps the anomalies: in this case it is much easier to wrongly separate normal points w.r.t. anomalies, indeed the first iteration of the model corresponding to the unsupervised training gives the highest anomaly score to the normal bottom left point of the toroid. On the contrary, at the 25-\textit{th} iteration the model has learnt the correct function and is able to perfectly classify all the points. This is also visible in the panels of the second row of Figure \ref{fig:toroid}, where the performance of the detector at each iteration is depicted with lines having different colors. As the model is allowed to query new points, the average detection performance quickly improves.

We decided to test the model on a set of $18$ real data openly available \cite{Rayana,Dua:2019}.
Table \ref{dataset_used} presents the dataset used to test the performance of the proposed model. These come from different domains like medicine, industry and natural sciences and are characterised by various number of points as well as percentage of anomalous data. Defining the contamination as the ratio between the number of anomalies and total number of samples of the dataset, they are characterized by a wide feature range between $6$ to $100$ and a contamination percentage between $0.9\%$ and $36\%$. It is very important to highlight how most of these datasets were built: they are adaptations of multi-class classification datasets to the anomaly detection task where one of the classes is under sampled and labelled as outlier. This means that points considered outliers are not just points randomly scattered in the features space like general anomalies, but they live in specific positions of the space that can be hard to be defined without labels. In this context, weakly supervised methods prove their efficacy, starting from an unsupervised guess, and improving continuously as labels are included in the model.

\input{dataset_table}

We used the average precision score metric to measure the results obtained. Splitting equally the dataset in training and testing set, we carried on a number of $25$ queries, and the experiments were conducted $50$ independent times to study the performances distribution. 
The experiments were performed on equipment with Intel Core i7-6800K CPU and 32 GB RAM. The implementation of \approach is freely available online\footnote{\url{https://github.com/tombarba/ActiveLearningIsolationForest} }. 

\begin{figure*}
   \centering
    \includegraphics[width=\textwidth]{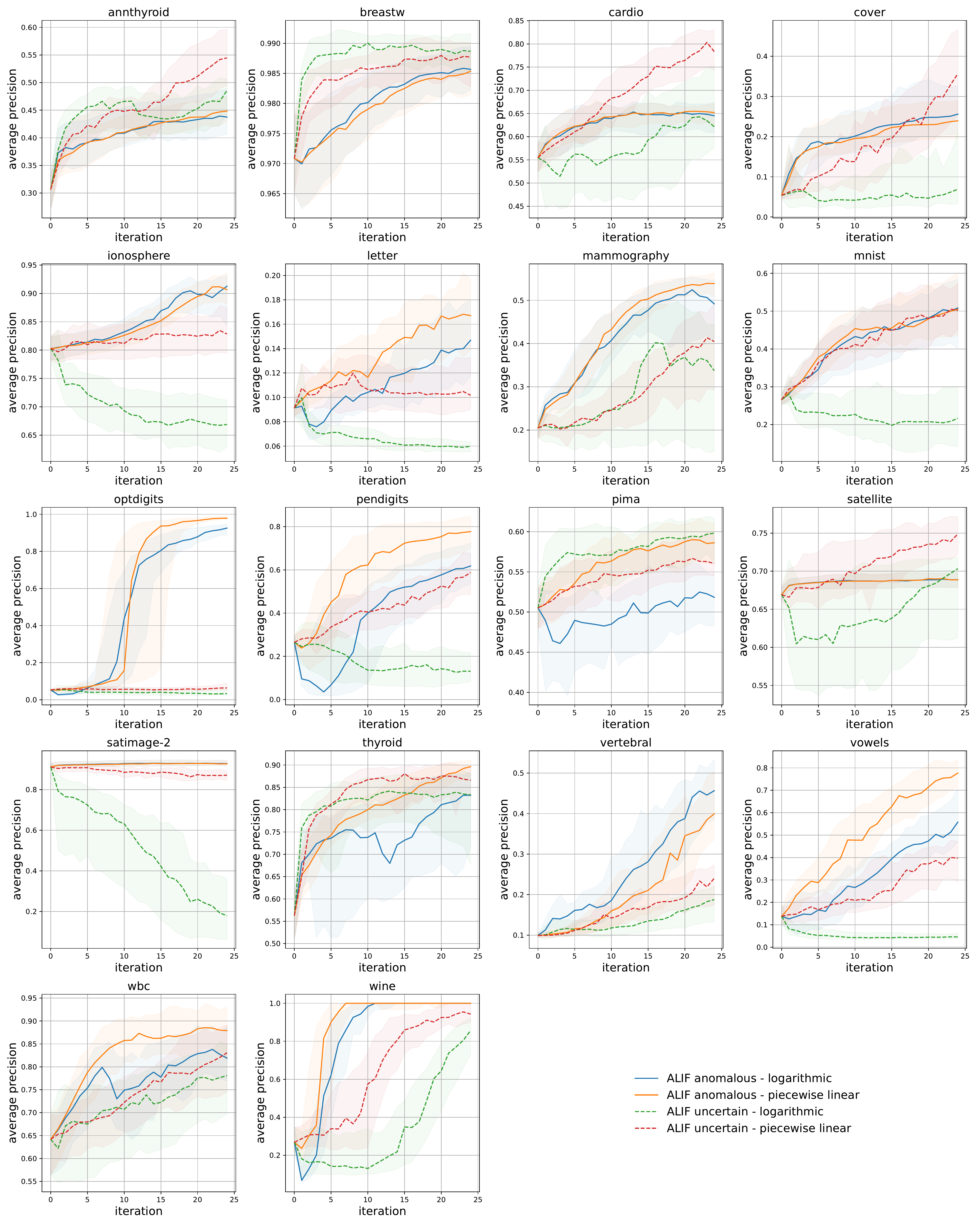}
    \caption{Performance of the four combinations of the proposed strategies with the datasets described in Table \ref{dataset_used}. As a general result, it can be noticed that querying the most anomalous point represents the most appropriate choice, overall leading to quicker improvements of the performance. When it comes to the update strategies, both the linear and the logarithmic depths seem to represent a reasonable choice. However, the linear depth appears to moderately be more stable.}
    \label{risultati_1}
\end{figure*}

First, we tested the four possible combinations of the two proposed update strategies together with the two query strategies so as to determine the best combination with respect to the majority of the considered test sets. Figure \ref{risultati_1} shows the obtained results.

Given that the first point of the curve represents the performance of the fully unsupervised Isolation Forest, it can be observed as, broadly speaking, all four proposed matches represent an improvement with respect to the performance of the Isolation Forest. The two trials of the "most anomalous" query are depicted in solid line, while the "maximum uncertainty" in dashed line. Even if there are some exceptions, in most cases the first policy seems the one having the fastest improvements. This is a very interesting aspect since the "most anomalous" strategy is the cheapest and most natural policy among the two considering the DSS scenario previously described. Concerning the updating strategy, as expected, the piece-wise linear is the one having the most stable improvements in the dataset and among the datasets. As a direct consequence, we decided to use the combination "most anomalous - piecewise linear" for the comparison to the other baseline and state-of-art competitor.

As our approach \approach essentially relies on designing an active learning based modification of the classical Isolation Forest, we decided to compare it with other tree based models: a fully supervised model, i.e. the Random Forest, and a weakly supervised model, the Isolation Forest - Active Anomaly Detection (IF-AAD).
The RF represents the baseline since it is the most well known but simple classification approach; unfortunately it requires samples from both the classes inlier/outlier and it is computationally expensive as every time the user labels a new data point the forest is retrained. As the RF does not have a default method to query its points, in the following comparison the points given to the RF for training are the points queried by our \approach.
On the other side IF-AAD is a active semi-supervised model  that does not need both class labels and is available online\footnote{\url{https://github.com/shubhomoydas/ad_examples}}. 

\begin{figure}
    \centering
    \includegraphics[width=\textwidth]{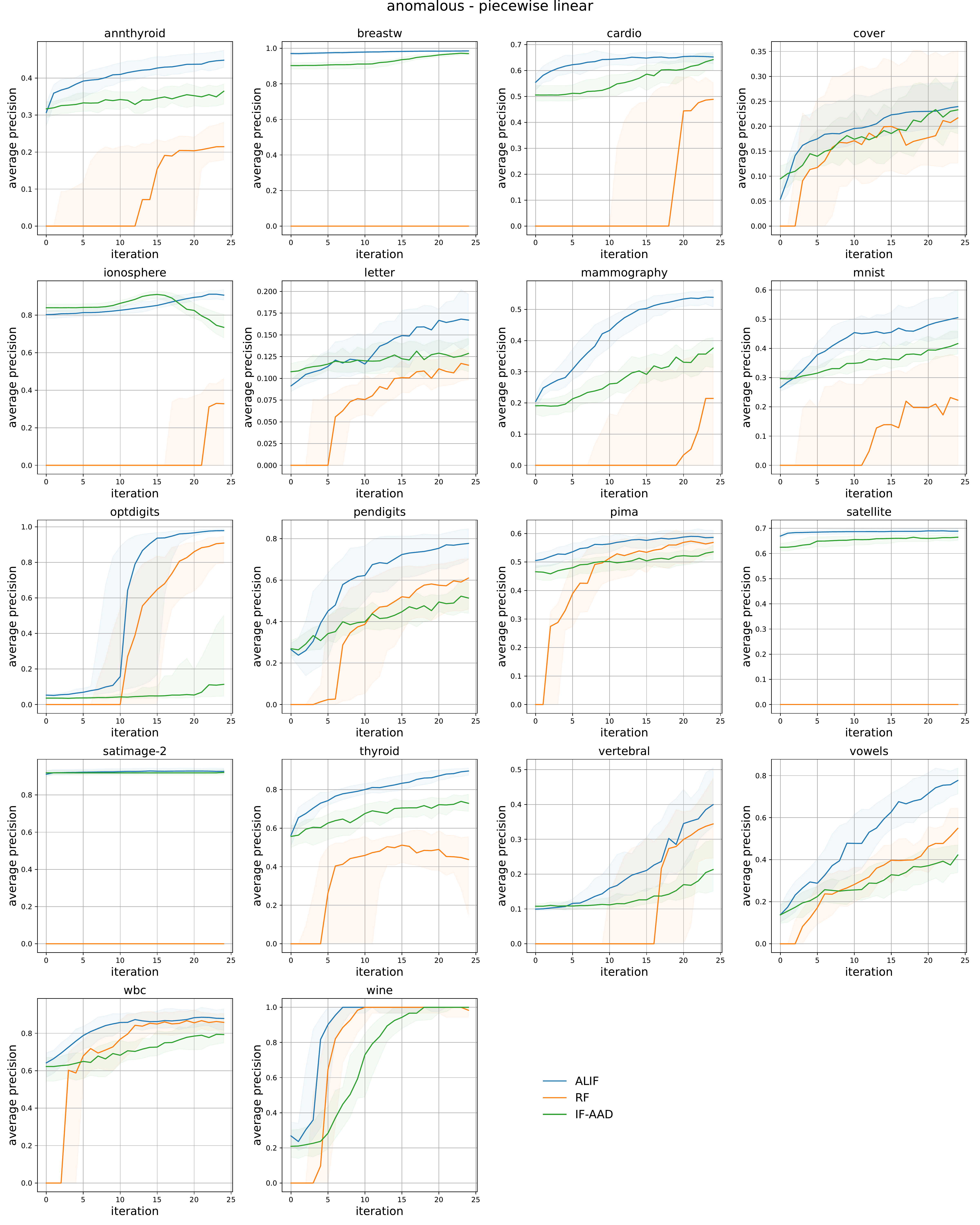}
    \caption{Comparison of most anomalous - piece-wise linear \approach with IF-AAD and RF. It can be observed that, in general our method represents the best course of action, having the highest performance score usually with a very small amount of labeled data.}
    \label{benchmark}
\end{figure}

Figure \ref{benchmark} and Table \ref{table:final_results} show the benchmark results: it is prominent that \approach obtains the finest results with a very little number of labels, generally outperforming IF-AAD. Due to the strong imbalance between inliers and outliers RF receives quite late both labels from the two classes, leading to poor detection performances: in \emph{breastw}, \emph{satellite}, and \emph{satimage-2} the querying process never got them over 25 iterations and 50 independent repetitions 

\begin{table}[]
\centering
\input{final_table}
\caption{Summary of the obtained results. The reported performances are the mean performance along the 25 iterations and 50 repetitions of the algorithm. \approach consistently beats the other tested approached on all the datasets, in particular the combination of the most anomalous query policy and the piece-wise linear leaf update.}
\label{table:final_results}
\end{table}



%% file: dataset_table.tex
\begin{table*}[]
\centering
	\begin{tabular}{@{}lccc c@{}} \hline
		\textbf{Dataset} & \textbf{Instances} & \textbf{Features} & \textbf{Anomalies}  & \textbf{Contamination}\\ \hline
		AnnThyroid   & 7200                & 6                  & 534        & 7.42\%        \\ 
		Breastw      & 683                 & 9                  & 239        & 35 \%        \\
		Cardio &	1831 & 21 & 176 & 9.6\% \\
		Cover & 286048 & 10 & 2747 & 0.9\% \\
		Ionosphere   & 351                 & 33                 & 126          &      36\%\\
		Letter & 1600 &  32 & 100 & 6.25\% \\
		Mammography  & 11183               & 6                  & 260           &  2.32\%   \\
		Mnist & 7603 & 100 & 700 & 9.2\% \\
		Optdigits	& 5216 & 64 &	150 & 3\% \\
		Pendigits    & 6870                & 16                 & 156          &  2.27\%    \\
		Pima         & 768                 & 8                  & 268          &  35\%    \\
		Satellite & 6435 & 36 & 2036  & 32\% \\
		Satimage-2	&5803&36&	71& 1.2\% \\
		Thyroid	& 3772 & 6 & 93 & 2.5\% \\
		Vertebral	&240&	6	&30 &12.5\% \\
		Vowels& 	1456&12	&50 &3.4\%   \\
		WBC	&278	&30	&21 &5.6\%   \\ 
		Wine & 129	& 13	& 10 & 7.7\% \\ \hline
		
	\end{tabular}
	\caption{Set of data used in the experimental phase. The first column gives the name of the dataset; the second column describes the number of instances contained in each set; the third column defines the total amount of features; the fourth column gives the number of outliers; the last column presents the contamination rates.}
	\label{dataset_used}
	\end{table*}

%% file: final_table.tex
\begin{adjustbox}{angle=270}
\begin{tabular}{llllllllll}
\toprule
{} &      {}  & \multicolumn{2}{l}{anom-log} & \multicolumn{2}{l}{anom-lin} & \multicolumn{2}{l}{unc-log} & \multicolumn{2}{l}{unc-lin} \\
{} & IF-AAD &       RF &   ALIF &       RF &   ALIF &      RF &   ALIF &      RF &   ALIF \\
\midrule
\textbf{annthyroid } &   0.34 &     0.11 &  0.41 &     0.11 &  0.41 &    0.11 &  0.44 &    0.22 &  \textbf{0.45} \\
\textbf{breastw    } &   0.93 &     0.00 &  0.98 &     0.00 &  0.98 &    0.28 &  \textbf{0.99} &    0.48 &  0.98 \\
\textbf{cardio     } &   0.56 &     0.15 &  0.63 &     0.16 &  0.63 &    0.34 &  0.57 &    0.49 &  \textbf{0.69} \\
\textbf{cover      } &   0.18 &     0.20 &  0.21 &     0.18 &  0.20 &    0.25 &  0.09 &    \textbf{0.46} &  0.19 \\
\textbf{ionosphere } &   0.84 &     0.08 &  \textbf{0.85} &     0.08 &  \textbf{0.85} &    0.47 &  0.70 &    0.52 &  0.82 \\
\textbf{letter     } &   0.12 &     0.08 &  0.11 &     0.07 &  \textbf{0.14} &    0.06 &  0.07 &    0.07 &  0.11 \\
\textbf{mammography} &   0.28 &     0.09 &  0.40 &     0.10 &  \textbf{0.43} &    0.04 &  0.28 &    0.18 &  0.28 \\
\textbf{mnist      } &   0.36 &     0.15 &  0.41 &     0.15 &  \textbf{0.43} &    0.25 &  0.24 &    0.42 &  0.42 \\
\textbf{optdigits  } &   0.11 &     0.46 &  0.49 &     0.38 &  \textbf{0.51} &    0.00 &  0.05 &    0.00 &  0.06 \\
\textbf{pendigits  } &   0.42 &     0.28 &  0.39 &     0.34 &  \textbf{0.59} &    0.10 &  0.19 &    0.24 &  0.42 \\
\textbf{pima       } &   0.50 &     0.44 &  0.49 &     0.43 &  0.56 &    0.34 &  \textbf{0.57} &    0.43 &  0.54 \\
\textbf{satellite  } &   0.65 &     0.01 &  0.69 &     0.01 &  0.68 &    0.42 &  0.64 &    0.49 &  \textbf{0.70} \\
\textbf{satimage-2 } &   0.92 &     0.00 &  \textbf{0.93} &     0.00 &  \textbf{0.93} &    0.25 &  0.48 &    0.53 &  0.88 \\
\textbf{thyroid    } &   0.66 &     0.37 &  0.69 &     0.33 &  \textbf{0.80} &    0.10 &  0.73 &    0.33 &  \textbf{0.80} \\
\textbf{vertebral  } &   0.14 &     0.16 &  \textbf{0.27} &     0.12 &  0.22 &    0.17 &  0.14 &    0.22 &  0.17 \\
\textbf{vowels     } &   0.29 &     0.29 &  0.33 &     0.30 &  \textbf{0.51} &    0.07 &  0.06 &    0.18 &  0.27 \\
\textbf{wbc        } &   0.71 &     0.68 &  0.76 &     0.68 &  \textbf{0.82} &    0.11 &  0.69 &    0.21 &  0.73 \\
\textbf{wine       } &   0.69 &     0.73 &  0.80 &     0.75 &  \textbf{0.85} &    0.28 &  0.38 &    0.47 &  0.64 \\
\bottomrule
\end{tabular}
\end{adjustbox}

%% file: conclusions.tex
\section{Conclusions}
\label{conclusions}

In this paper a modification of the unsupervised Isolation Forest named \approach has been suggested to improve the performance of the standard algorithm.
Due to the lack of fully labelled datasets, Anomaly Detection is often performed by means of unsupervised models. However, as anomalies heavily depend on the context and are very domain specific, unsupervised models may be unable to detect them because of different definition of anomaly.
To solve this problem we suggest a model that starting from an unsupervised model, iteratively tunes the model towards the user-definition of anomaly. This allows to enhance the detection performances, avoiding the need to fully label the training dataset and keeping as low as possible the number of required labels. Indeed this approach  takes advantage of the Active Learning framework where the model is able to query the user and select the most interesting samples to label.
\approach relies on two important and inter-connected steps: the query policy that selects the point to be labelled, and the model update policy that allows the model to actually learn from the new query. 
From experiments performed on real datasets  it turned out it is better to ask to label the most anomalous point, leading to a cheap and natural query strategy in practice. Concerning the update policy, the model does not need to fully retrain the forest, but needs just a simple update of the leaf depth, with a cost $O(n_T)$. Regarding the query strategy, the required cost is $O(n_X n_T)$. The cheap time complexity together with the fact that \approach does not need labels of both anomalies and normal points provides a great advantage compared to other state-of-art algorithms.
Comparing \approach with other methods, it turns out it is also generally much faster in the learning of the correct anomaly definition, when the labelling effort needs to be kept low.

Future works will investigate the different query and updating strategies: the information of the number of data points in each leaf might be included in the algorithm, to estimate the uncertainty on the leaf depth correction. Moreover, hybrid query strategies that use a combination of the previously described, might lead to even better results.